\documentclass[aip,graphicx]{revtex4-1}

\usepackage[utf8]{inputenc}
\usepackage{xcolor}
\usepackage{hyperref}
\usepackage{amssymb}
\usepackage{amsmath}
\usepackage{lineno}
\usepackage{array}
\usepackage{caption}
\usepackage{subcaption}
\usepackage{multirow}
\usepackage{csquotes}
\usepackage{amsfonts}
\usepackage{relsize}
\usepackage{bbm}
\usepackage{comment}
\usepackage{graphicx}
\usepackage{makecell}
\usepackage{amsmath}

\draft 

\begin{document}

\preprint{Version 1}

\title{Benchmarking YOLOv5 and YOLOv7 models with DeepSORT for droplet tracking applications} 



\author{Mihir Durve}

\email{mihir.durve@iit.it}
\affiliation{Center for Life Nano- \& Neuro-Science, Fondazione Istituto Italiano di Tecnologia (IIT), viale Regina Elena 295, 00161 Rome, Italy}

\author{Sibilla Orsini}
\affiliation{NEST, Istituto Nanoscienze-CNR and Scuola Normale Superiore, Piazza San Silvestro 12, Pisa, 56127,  Italy}
\affiliation{Istituto per le Applicazioni del Calcolo del Consiglio Nazionale delle Ricerche, via dei Taurini 19, Roma, 00185, Italy}

\author{Adriano Tiribocchi}
\affiliation{Istituto per le Applicazioni del Calcolo del Consiglio Nazionale delle Ricerche, via dei Taurini 19, Roma, 00185, Italy}

\author{Andrea Montessori}
\affiliation{Dipartimento di Ingegneria, Università degli Studi Roma tre, via Vito Volterra 62, Rome, 00146, Italy}

\author{Jean-Michel Tucny}
\affiliation{Center for Life Nano- \& Neuro-Science, Fondazione Istituto Italiano di Tecnologia (IIT), viale Regina Elena 295, 00161 Rome, Italy}
\affiliation{Dipartimento di Ingegneria, Università degli Studi Roma tre, via Vito Volterra 62, Rome, 00146, Italy}

\author{Marco Lauricella}
\affiliation{Istituto per le Applicazioni del Calcolo del Consiglio Nazionale delle Ricerche, via dei Taurini 19, Roma, 00185, Italy}

\author{Andrea Camposeo}
\affiliation{NEST, Istituto Nanoscienze-CNR and Scuola Normale Superiore, Piazza San Silvestro 12, Pisa, 56127,  Italy}

\author{Dario Pisignano}
\affiliation{NEST, Istituto Nanoscienze-CNR and Scuola Normale Superiore, Piazza San Silvestro 12, Pisa, 56127,  Italy}
\affiliation{ Dipartimento di Fisica, Università di Pisa, Largo B. Pontecorvo 3, Pisa, 56127, Italy}

\author{Sauro Succi}
\affiliation{Center for Life Nano- \& Neuro-Science, Fondazione Istituto Italiano di Tecnologia (IIT), viale Regina Elena 295, 00161 Rome, Italy}
\affiliation{Department of Physics, Harvard University, 17 Oxford St, Cambridge, MA 02138, United States}


\date{\today}

\begin{abstract}
Tracking droplets in microfluidics is a challenging task. The difficulty arises in choosing a tool to analyze general microfluidic videos to infer physical quantities. The state-of-the-art object detector algorithm \textit{You Only Look Once (YOLO)}  and the object tracking algorithm \textit{Simple Online and Realtime Tracking with a Deep Association Metric  (DeepSORT)} are customizable for droplet identification and tracking. The customization includes training YOLO and DeepSORT networks to identify and track the objects of interest. We trained several YOLOv5 and YOLOv7 models and the DeepSORT network for droplet identification and tracking from microfluidic experimental videos. We compare the performance of the droplet tracking applications with YOLOv5 and YOLOv7 in terms of training time and time to analyze a given video across various hardware configurations. Despite the latest YOLOv7 being 10\% faster, the real-time tracking is only achieved by lighter YOLO models on RTX 3070 Ti GPU machine due to additional significant droplet tracking costs arising from the DeepSORT algorithm. This work is a benchmark study for the YOLOv5 and YOLOv7 networks with DeepSORT in terms of the training time and inference time for a custom dataset of microfluidic droplets.
\end{abstract}


\maketitle 


\section{Introduction}
\label{intro}
A subset of machine learning-based tools, called computer vision tools, deal with object identification, classification, and tracking in images or videos. State-of-the-art computer vision tools can read handwritten text~\cite{kang2022,coquenet,darmatasia,ahlawat}, find objects in images \cite{zou2019object,Joseph_2021_CVPR,brownlee,zou_arxiv}, find product defects \cite{prabhu,JOHNRAJAN2021}, make a medical diagnosis from medical images with accuracy surpassing humans \cite{Esteva2021,Bhargava2021} and object tracking \cite{Soleimanitaleb,Xu2021}, just to name a few. In the last few years, they have been increasingly consolidating their place in all scientific fields and industries as reliable and fast analysis methods. 

Computer vision tools have shown remarkable success in studying microfluidic systems. Artificial neural networks, for example, can predict physical observables, such as flow rate, chemical composition, etc., from images of microfluidics systems with high accuracy, thus reducing hardware requirements to measure these quantities in an microfluidics experiment \cite{hadikhani,mahdi}. More recently, a convolutional autoencoder model was trained to predict stable vs unstable droplets from their
shapes within a concentrated emulsion \cite{khor}.

Another application of computer vision tools in microfluidics is tracking droplets in experiments and simulation studies, such as the ones in Ref. \cite{bogdan2022,benzi1992lattice,montessori2019modeling,montessori2021_2}. Droplet recognition and tracking can yield rich information without needing human intervention. For example, counting droplet numbers, measuring flow rate, observing droplets size distribution and computing statistical quantities are cumbersome to measure with the manual marking of the droplets across several frames. Two natural questions, while using computer vision tools for image analysis, are i) how accurate the application is in terms of finding and tracking the objects, and ii) how fast the application is in analyzing each image. A typical digital camera operates at 30 frames per second (fps), thus one challenge 
is to analyze the images at the same or higher rate for real-time applications. 

Along with a few other algorithms, You Only Look Once (YOLO) has the capability to analyze images at a few hundred frames per second \cite{redmon,redmon1} and is designed to detect 80 classes of objects in a given image. The very first version of YOLO was introduced back in 2015 and the subsequent versions have been focused on making the algorithm faster and more accurate at detecting objects. The latest release of YOLO is its 7th version \cite{yolov7}, with a reported significant gain in speed and accuracy for object detection in standard datasets containing several objects in realistic scenes. 
In our previous study, we trained YOLO version 5 and DeepSORT for real-time droplet identification and tracking in microfluidic experiments and simulations \cite{durve_pof,durve_epjp}, and we reported the image analysis speed 
for various YOLOv5 models. In this one, we train the latest YOLOv7 models along with DeepSORT and compare performance and image analysis speed of these models with the previous one. In particular, this paper studies and compares training time, droplet detection accuracy and inference time for an application that combines YOLOv5/YOLOv7 with DeepSORT for droplet recognition and tracking. 

\section{Experimental methods }

\begin{figure}
\centering
  \includegraphics[width=0.8\textwidth]{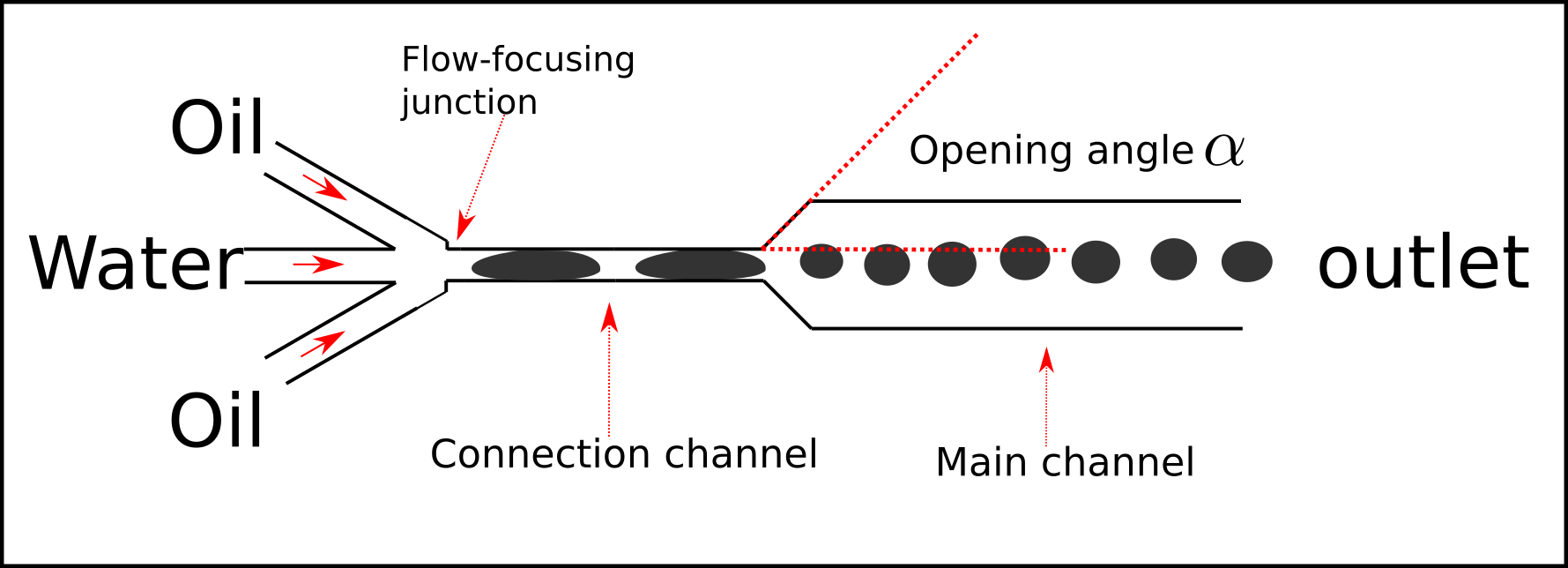}
\caption{Schematic representation of the microfluidic device used for the droplet generation.}
\label{fig_setup}       
\end{figure}

The images analyzed in this study were obtained from a microfluidic device for the generation of droplets exploiting a flow-focusing configuration (scheme of the device in Fig.\ref{fig_setup}). The device has two inlets for oil flow, one inlet for the flow of an aqueous solution, a Y-shaped junction for droplet generation and an expansion channel. The latter is connected to an outlet for collecting the two-phase emulsion. The device was realized by using a stereolithography system (Enviontec, Micro Plus HD) and the E-shell® 600 (Envisiontec) as pre-polymer. The continuous phase consists of silicone oil (Sigma Aldrich, oil viscosity 350 cSt at 25°C), while an aqueous solution constitutes the dispersed phase. The latter was made by dissolving 7 mg of a black pigment (Sigma Aldrich, Brilliant Black BN) in 1 mL of distilled water. Both phases were injected through the inlets at constant flow rates by a programmable syringe pump with two independent channels (Harvard Apparatus, model 33). The images analyzed in this study were obtained by using a flow rate of 10 $\mu$l/min and 150 $\mu$l/min for the dispersed phase and the continuous phase, respectively. The droplet formation is imaged by using a stereomicroscope (Leica, MZ 16 FA) and a camera (Photron, fastcam APX RS). The fast camera acquired the images at 3000 frames per second (fps). This image capture rate is far higher than any present algorithm's real-time object detection capabilities. The image playback rate is to 30 fps. The sequences of images were stored as AVI video files. Later, images from the video were used to train YOLO and DeepSORT models as described in the following section. 
\label{exp_details}

\section{Training YOLOv5 and YOLOv7 models}

\begin{figure}
\centering

  \includegraphics[width=0.8\textwidth]{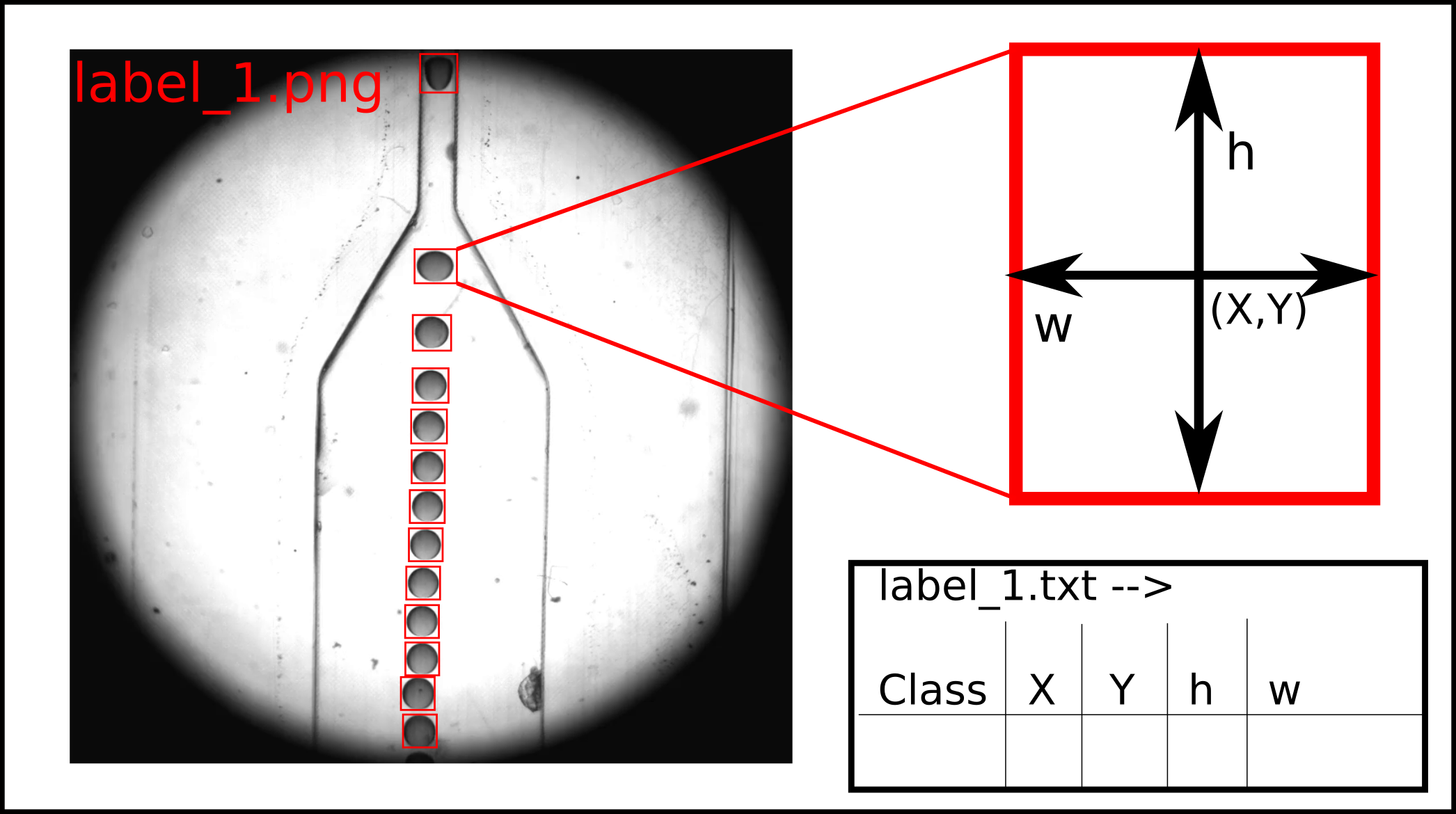}
\caption{Example from custom training dataset to train YOLOv5 and YOLOv7 object detector models. Each object is manually placed in a rectangle (called the bounding box) and the dimensions of the rectangle are noted in an associated label file.}
\label{fig_td}       
\end{figure}
The steps required to train YOLOv5 and YOLOv7 are identical. First, a training dataset is prepared by manually annotating 1000 images taken from a microfluidics experiment as described in Sec \ref{exp_details}. Each image in this dataset has approximately 13 to 14 droplets. One example from the training dataset is shown in Fig.\ref{fig_td}. The droplets in these images are identified, and the dimensions of a rectangle that fully covers the droplet are noted in a separate text file called the label file. We used PyTorch implementation of YOLOv5 \cite{yolov5_git} and YOLOv7 \cite{yolov7_git} to train several YOLO models on an HPC system on a single node containing two Intel(R) Xeon(R) Gold 6240 CPU @ 2.60GHz Cascade Lake and NVIDIA Tesla V100 GPU with 32 GB VRAM. A typical training time is mentioned in Table \ref{tab_tt}.

\begin{table*}
\centering

\caption{YOLO models training time on the same machine with an identical training dataset. The YOLO model descriptions can be found in Ref.\cite{yolov5_git} for v5 and in Ref.\cite{yolov7} for v7 }
\label{tab_tt}       
\begin{tabular}{llllll}
\hline\noalign{\smallskip}
Model & Parameters & Image size & Epoch & Total time & Time per epoch \\
 & (millions) & (pixels) & & (sec) &  (sec) \\
\noalign{\smallskip}\hline\noalign{\smallskip}
\noalign{\smallskip}\hline\noalign{\smallskip}

YOLOv5s & 7.2  &  640 & 1000 & 15465.6 & 15.465  \\
YOLOv5m & 21.2 & 640 & 1000 & 27075.6 & 27.075\\
YOLOv5l	& 46.5 & 640 & 1000 & 39420.2 & 39.420 \\
YOLOv5x & 86.7 & 640 & 1000 & 50670.0 & 50.670 \\
YOLOv5s6 & 12.6 & 1280 & 1000 & 17456.4 & 17.456 \\
YOLOv5m6 & 35.7 & 1280 & 1000 & 30355.2 & 30.355 \\
YOLOv5l6 & 76.8 & 1280 & 1000 & 44521.2 & 44.521 \\
YOLOv5x6 & 140.7 & 1280 & 1000 & 54630.0 &  54.630\\

\noalign{\smallskip}\hline\noalign{\smallskip}
YOLOv7-tiny & 6.2  & 640 & 1000 & 14423.6 & 14.420 \\
YOLOv7 & 36.9  & 640 & 1000 & 26740.8 & 26.740 \\
YOLOv7-x & 71.3  & 640 & 1000 & 45231.4 & 45.230 \\
YOLOv7-w6 & 70.04 & 1280 & 1000 & 66740.4  & 66.740 \\
YOLOv7-e6 & 97.2 & 1280 & 1000 & 100399.5 & 100.399 \\
YOLOv7-d6 & 154.7 & 1280 & 1000 & 140428.2 & 140.428 \\
YOLOv7-e6e & 151.7 & 1280 & 1000 & 135014.2 & 135.010 \\
\noalign{\smallskip}\hline
\end{tabular}
\end{table*}

\begin{figure*}
  \includegraphics[width=1.0\textwidth]{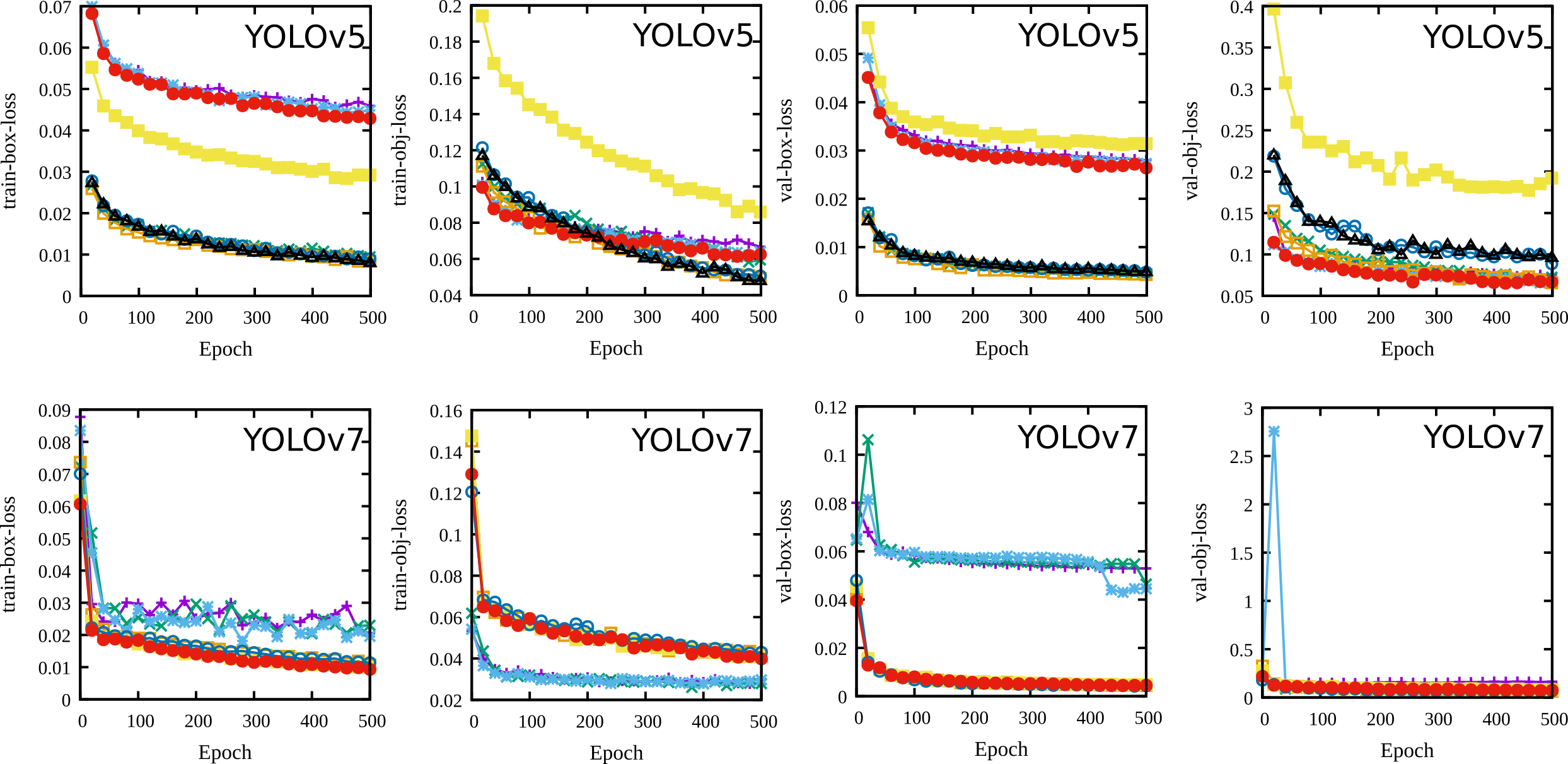}
\caption{Loss function during the YOLOv5 and YOLOv7 as the training progress. Figure legends are the same as in Fig. \ref{fig_map}.}
\label{fig_loss}       
\end{figure*}

During the training phase, a subset of data (called a batch) is passed through the network and a loss value is computed using the difference between the network's predictions and the ground truth provided in the label file. The loss value is then used to update the network's trainable parameter to minimize the loss in subsequent passes. An epoch is said to be completed when all of the training data is passed through the network. YOLO's loss calculation takes into account the error in bounding box prediction, error in object detection, and error in object classification \cite{redmon}. The loss value components computed with training and validation data are shown in Fig \ref{fig_loss}.   

\begin{figure}
\centering

  \includegraphics[width=0.8\textwidth]{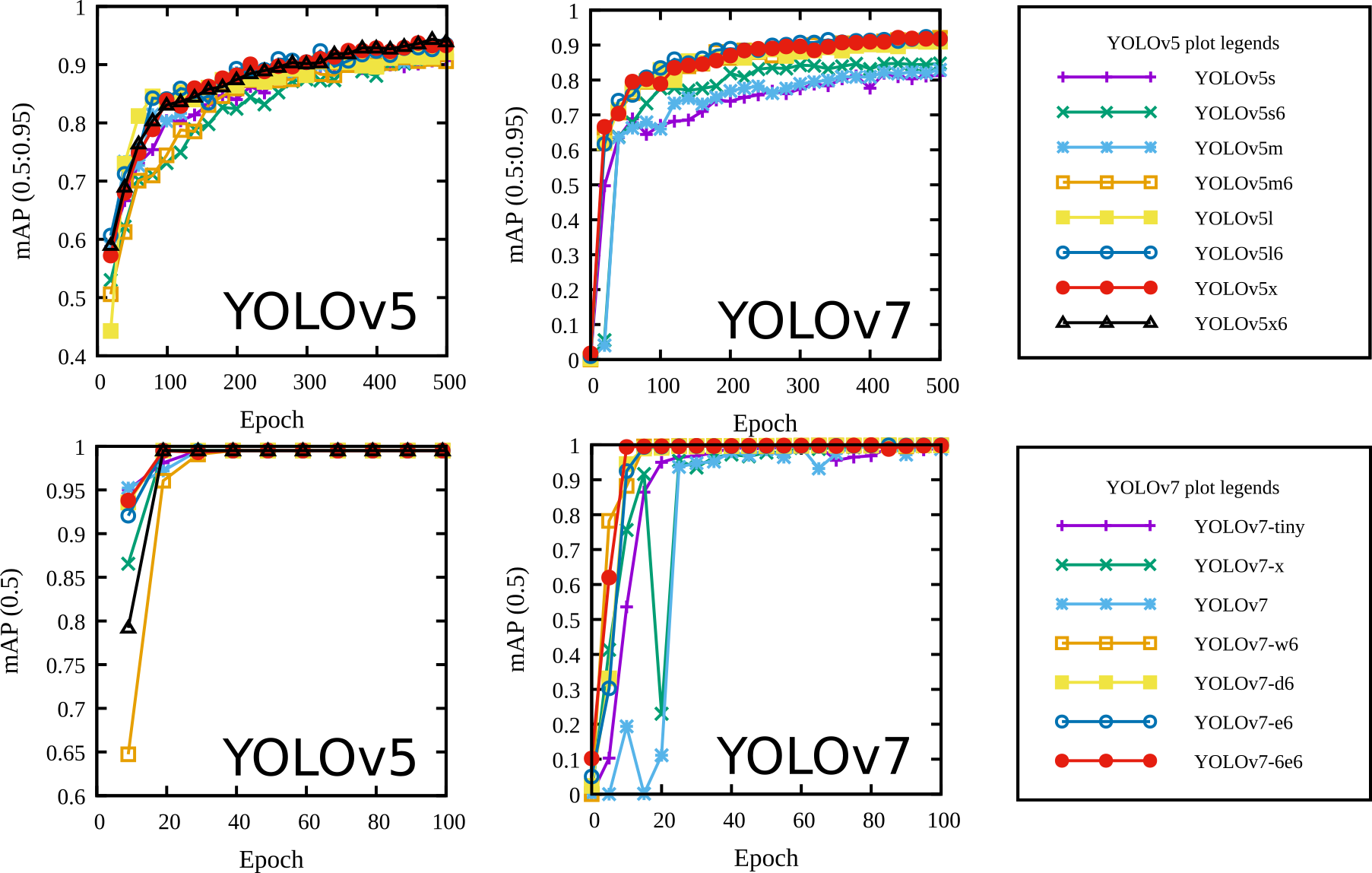}
\caption{Mean Average Precision (mAP) comparison between YOLOv5 and YOLOv7 models with custom dataset}
\label{fig_map}       
\end{figure}

During the training phase, the quality of YOLOv5 and YOLOv7 models is measured with a well-known mean average precision (mAP), which is calculated with an Intersection over Union (IoU) threshold of 0.5 (see Fig. \ref{fig_map}). For both versions, mAP value quickly saturates to unity after training with 20 epochs. Similarly, the average of mAP calculated with IoU threshold of 0.5 to 0.95 in steps of 0.05 shows no significant differences between YOLOv5 and YOLOv7 models. 

\section{Inference with YOLO and DeepSORT}
After the models are trained, they can be deployed for real-world applications. One challenging milestone for any computer vision application is to use it in real time, i.e. when the image analysis speed exceeds 30 fps. YOLO models on their own do deliver real-time performance. In table \ref{tab_CPU} and \ref{tab_GPU}, we show the total time for droplet identification and tracking, combining YOLOv5/YOLOv7 with DeepSORT on two hardware configurations. The benchmarking study was carried out on an MSI G77 Stealth laptop with i7-12700H, 32 GB RAM, and NVIDIA RTX 3070 Ti 8 GB VRAM GPU. Running on GPU, we observe approximately 10\% improvement in the inference speed for YOLOv7 over YOLOv5. However, additional time by the object tracking algorithm DeepSORT is comparable with havier YOLO models. The single application combining object identification and tracking can deliver real-time tracking with lighter YOLO models, but they fall below the real-time tracking mark with heavier YOLO models. 
Finally, a video of droplet tracking is provided in supplemental material (see SM1.avi). 
%
\begin{table}
\centering

\caption{Inference time per frame - CPU}
\label{tab_CPU}       
\begin{tabular}{lllll}
\hline\noalign{\smallskip}
Model & YOLO & DeepSORT & Total time & FPS \\
 & (sec) & (sec) & (sec) &   \\
\noalign{\smallskip}\hline\noalign{\smallskip}
\noalign{\smallskip}\hline\noalign{\smallskip}

YOLOv5s & 0.1097	 & 	0.1207	 & 	0.2304	 & 	4.34 \\
YOLOv5m & 0.1213	 & 	0.1206	 & 	0.2419	 & 	4.13 \\
YOLOv5l & 0.3494	 & 	0.1197	 & 	0.4690	 & 	2.13 \\
YOLOv5x & 0.5925	 & 	0.1142	 & 	0.7067	 & 	1.42 \\
YOLOv5s6 & 0.2719	 & 	0.1175	 & 	0.3894	 & 	2.57 \\
YOLOv5m6 & 0.6507	 & 	0.1131	 & 	0.7638	 & 	1.31 \\
YOLOv5l6 & 1.2471	 & 	0.1188	 & 	1.3659	 & 	0.73 \\
YOLOv5x6 & 2.1758	 & 	0.1283	 & 	2.3041	 & 	0.43 \\

\noalign{\smallskip}\hline\noalign{\smallskip}
YOLOv7-tiny &  0.1064	 & 	0.1175	 & 	0.2240	 & 	4.47 \\
YOLOv7-x &  0.1152	 & 	0.1184	 & 	0.2336	 & 	4.28 \\
YOLOv7 &  0.3564	 & 	0.1143	 & 	0.4706	 & 	2.12 \\
YOLOv7-w6 &  0.2583	 & 	0.1199	 & 	0.3781	 & 	2.64 \\
YOLOv7-e6 &  0.5528	 & 	0.1131	 & 	0.6658	 & 	1.50 \\
YOLOv7-d6 &  1.2035	 & 	1.3107	 & 	2.5142	 & 	0.40 \\
YOLOv7-e6e &  0.6312	 & 	0.1188	 & 	0.7500	 & 	1.33 \\

\noalign{\smallskip}\hline
\end{tabular}
\end{table}

\begin{table}
\centering

\caption{Inference time per frame - GPU}
\label{tab_GPU}       
\begin{tabular}{lllll}
\hline\noalign{\smallskip}
Model & YOLO & DeepSORT & Total time & FPS \\
 & (sec) & (sec) & (sec) &   \\
\noalign{\smallskip}\hline\noalign{\smallskip}
\noalign{\smallskip}\hline\noalign{\smallskip}

YOLOv5s & 0.0057  & 0.0235 & 0.0292 & 34.27 \\
YOLOv5m & 0.0076 &  0.0156 &  0.0232  & 43.05 \\
YOLOv5l & 0.0192 &  0.0258 &  0.0450  & 22.24 \\
YOLOv5x & 0.0196 &  0.0229 &  0.0425  & 23.53 \\
YOLOv5s6 &  0.0162 &  0.0256 &  0.0418  & 23.92 \\
YOLOv5m6 & 0.0261 &  0.0304 &  0.0565  & 17.70 \\
YOLOv5l6 &0.0384 &  0.0237 &  0.0621  & 16.11  \\
YOLOv5x6 &0.0696 &  0.0186 &  0.0881  & 11.34\\

\noalign{\smallskip}\hline\noalign{\smallskip}

YOLOv7-tiny & 0.0049	 & 	0.0241	 & 	0.0290	 & 	34.48 \\
YOLOv7-x & 0.0065	 & 	0.0244	 & 	0.0309	 & 	32.40 \\
YOLOv7 & 0.0175	 & 	0.0217	 & 	0.0392	 & 	25.53 \\
YOLOv7-w6 & 0.0176	 & 	0.0221	 & 	0.0397	 & 	25.16 \\
YOLOv7-e6 & 0.0138	 & 	0.0256	 & 	0.0394	 & 	25.35 \\
YOLOv7-d6 & X & 	X	 & 	 X	 & 	X \\
YOLOv7-e6e & X	 & 	X	 & 	X	 & 	X \\

\noalign{\smallskip}\hline
\end{tabular}
\end{table}

\section{Conclusion}

This paper studied two versions of YOLO object detector models coupled with DeepSORT tracking algorithms. The benchmarks studied in this work serve as a guide for computational resource requirements to train the networks and mention expected inference time for various models on diverse hardware configurations.

YOLOv5 and YOLOv7 networks were trained with identical training datasets on the same HPC machine with NVIDIA-V100 GPU. The training time per epoch  is comparable for lighter YOLOv5 and YOLOv7, but the heavier YOLOv7 models take almost double the time to complete the training.

We observe a significant increase in inference speed in YOLOv7 models compared to their YOLOv5 counterparts, as one would expect. Moreover, we report detailed computational costs on object detection and object tracking routines and the overall performance of the combined application. Lighter YOLO models are much quicker to identify objects in comparison with the time taken by DeepSORT to track them. However, the object identification time increases with the increasing complexity of the object-detecting networks. Thus, it is crucial to choose the right YOLO network and hardware configuration for real-time tracking at the cost of the bounding box accuracy.

\section{Acknowledgements}
The authors acknowledge funding from the European Research Council Grant Agreement No. 739964 (COPMAT) and ERC-PoC2 grant No. 101081171 (DropTrack). We gratefully acknowledge the HPC infrastructure and the Support Team at Fondazione Istituto Italiano di Tecnologia.

\bibliography{Ref}

\end{document}